\documentclass{article}
\usepackage[T1]{fontenc}
\usepackage{microtype}
\usepackage{graphicx}
\usepackage{booktabs}
\usepackage{amsmath,amssymb,amsthm}
\usepackage{hyperref}
\usepackage{xurl}
\ifdefined\tmlrcopy
\usepackage{tmlr}
\else
\ifdefined\anonymouscopy
\usepackage{icml2026}
\else
\usepackage[preprint]{icml2026}
\fi
\fi
\usepackage{enumitem}
\usepackage{multirow}
\usepackage{xcolor}
\definecolor{inkblue}{HTML}{234A75}
\hypersetup{colorlinks=true,citecolor=inkblue,urlcolor=inkblue,linkcolor=inkblue}

\newcommand{\E}{\mathbb{E}}
\newcommand{\R}{\mathbb{R}}

\ifdefined\tmlrcopy
\title{A Controlled Audit of Personal AI Memory\\for Rating Prediction}
\author{Anonymous authors}
\else
\icmltitlerunning{A Controlled Audit of Personal AI Memory for Rating Prediction}
\fi
\newcommand{\CoatTargets}{2,960}

\begin{document}
\ifdefined\tmlrcopy
\maketitle
\else
\twocolumn[
\icmltitle{A Controlled Audit of Personal AI Memory\\for Rating Prediction}
\begin{icmlauthorlist}
\ifdefined\anonymouscopy
\icmlauthor{Anonymous Author}{anon}
\else
\icmlauthor{Shivam Gupta}{personal}
\fi
\end{icmlauthorlist}
\ifdefined\anonymouscopy
\icmlaffiliation{anon}{Affiliation withheld in this review copy}
\icmlcorrespondingauthor{Anonymous Author}{withheld}
\else
\icmlaffiliation{personal}{Independent research}
\icmlcorrespondingauthor{Shivam Gupta}{shivam1720406@gmail.com}
\fi
\icmlkeywords{personalization, agent memory, language models, evaluation, recommendation}
\vskip 0.3in
]
\printAffiliationsAndNotice{}
\fi
\hypersetup{pdftitle={A Controlled Audit of Personal AI Memory for Rating Prediction},pdfsubject={Research preprint. Not peer reviewed.}}
\ifdefined\anonymouscopy
\hypersetup{pdfauthor={Anonymous Authors},pdfsubject={Prepared anonymous manuscript. Not submitted or peer reviewed.}}
\fi
\begin{abstract}
In structured rating prediction, does a personal AI use historical item-rating associations, or mainly the user's rating tendencies? We audit this distinction by permuting historical ratings within each user while preserving the exact rating distribution, item support, and metadata. We combine this control with full history, native memory extraction, and matched numerical readers in a publicly frozen evaluation of 400 held-out user profiles and 6,160 target ratings across Coat and MovieLens. On Coat, the tested Qwen-written Mem0 pipeline increases user-macro mean absolute error relative to full history by 0.084 for Qwen and 0.149 for Phi; both family-adjusted bootstrap intervals exclude zero. Correct historical assignments help both readers on Coat, but the corresponding MovieLens effects are smaller and inconclusive after adjustment. A history-only ridge reader outperforms Qwen in both domains and Phi on MovieLens, while the Coat Phi comparison is unresolved. All 2,800 reader calls, including 150 invalid outputs, are retained under a fixed fallback rule. A separate implementation verifies inputs, metrics, and all ten primary contrasts. The contribution is a reproducible diagnostic study showing why extraction, association use, output reliability, and reader choice require separate evaluation.

\end{abstract}

\section{Introduction}
A personal AI should use past experience to make better decisions for a particular person. A memory system that lowers rating-prediction error appears to meet that requirement. Yet the improvement can arise from a much simpler behavior: learning that a person usually gives low or high ratings. This adaptation is useful, but it does not establish that the system knows which items that person prefers. The distinction is already recognized in recommendation evaluation \citep{tan2025preferences}. It becomes especially important when a memory writer transforms detailed experience into a natural-language profile and a separate model reads that profile.

There are three different places where this pipeline can succeed or fail. The source history may contain useful item-specific associations. The writer may preserve or distort those associations. The reader may then use the retained information effectively or fall back on broad rating tendencies. Comparing an extracted memory against no history confounds these stages. A fluent description can be useful even if it mainly preserves a user's rating level; conversely, a reader can underperform despite receiving the entire source history.

This paper studies these distinctions through controlled interventions on real, observed ratings. Our central control permutes rating values \emph{within the same user's history}, holding the rated items, their attributes, the history length, and the exact rating multiset fixed. The contrast asks whether the correct item-rating assignment improves prediction under a fixed reader. It does not replace a person's history with another person's, and it does not assert that all personalization has been removed. Together with a full-history comparison, a native memory writer, and matched statistical readers, this yields a diagnostic study of extraction and evidence use.

The empirical contribution is a publicly frozen evaluation on 200 users in each of two domains, held out from predictive development. Coat provides randomly elicited target ratings after self-selected histories; MovieLens provides a second domain with self-selected observed ratings and genre-only inputs. Two quantized open-weight readers are evaluated without tuning on the evaluation users. On Coat, both readers also receive the same memories produced by a pinned native Mem0 pipeline. We report every predeclared comparison, malformed outputs, statistical baselines with matching information, and construction costs. The paired design measures how these specific changes in source evidence, memory construction, and reader choice affect operational error.

The results separate failures that a single memory-versus-no-memory score would obscure. Native memory improves on no history on Coat, yet increases error relative to the source history for both readers. Correct item-rating assignments matter on Coat even when the exact rating distribution is preserved, whereas that benefit is not established on MovieLens after family adjustment. Numerical reading is also competitive: a history-only ridge model has lower error than three of the four full-history language-model conditions, with the remaining comparison unresolved.

Our contributions are: (i) an explicit set of input interventions and estimands for separating native extraction from historical-association use; (ii) a frozen, two-domain evaluation with matched-history baselines and complete failure accounting; and (iii) an auditable release with per-user measurements, independently implemented calculation checks, and resource accounting. These contributions are empirical and methodological in their application. They do not introduce a new permutation method, establish uniquely individual taste, or assert a universal ranking of memory systems.

\section{What the Controls Identify}
\begin{figure*}[t]
\centering
\includegraphics[width=.96\textwidth]{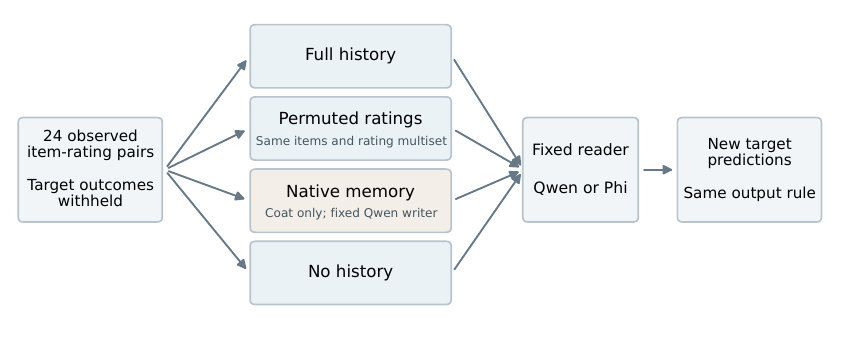}
\caption{The same source history supports three evidence controls and, on Coat, native memory extraction. Target outcomes never enter construction or prediction. Each reader receives the same new target list in every condition. A separate ridge reader uses the same observed history and metadata.}
\label{fig:design}
\end{figure*}
\subsection{Setting and interventions}
Figure~\ref{fig:design} summarizes the audit. For user $u$, let $H_u=\{(i_j,x_j,y_j)\}_{j=1}^{m}$ be the observed history, with item identifier $i_j$, metadata $x_j$, and rating $y_j\in[1,5]$. Let $Q_u$ contain the identifiers and metadata of new target items. Target outcomes $Y_u$ are withheld from every prediction prompt and memory write. A reader $f$, including its declared parser and fallback policy, returns a vector $f(C,Q_u)$ from evidence $C$. A writer $W$ constructs natural-language memory $W(H_u)$ before the targets are presented.

For a permutation $\pi_u$ of the $m$ historical positions, define
\begin{equation}
 H_u^{\pi}=\{(i_j,x_j,y_{\pi_u(j)})\}_{j=1}^{m}.
\end{equation}
The permutation preserves all rated items and their attributes, as well as the complete empirical distribution of the historical ratings. It changes only their assignment to items. Item order remains fixed. We compare $C=\varnothing,H_u,H_u^\pi$, and, on Coat, $W(H_u)$. The same target list, reader, decoding settings, and output parser apply to every condition.

For a target-prediction vector $z$, define $\ell_u(z)=|Q_u|^{-1}\sum_{i\in Q_u}|z_i-y_{ui}|$. The reader loss is
\begin{equation}
 L_u(C)=\ell_u(f(C,Q_u)).
\end{equation}
The three primary contrasts, averaged across evaluation users, are
\begin{align}
 A_f &= \E_u[L_u(H_u^\pi)-L_u(H_u)],\\
 X_f &= \E_u[L_u(W(H_u))-L_u(H_u)],\\
 N_f &= \E_u[L_u(H_u)-\ell_u(g(H_u,Q_u))].
\end{align}
Here $g$ is a fixed, development-selected history-only ridge reader. Positive $A_f$ means the correct assignments help this reader. Positive $X_f$ means the tested extraction pipeline increases error relative to the original history. Positive $N_f$ means the numerical reader has lower error. Only $A_f$ and $N_f$ are replicated in MovieLens; memory extraction is evaluated in Coat alone.

These are controlled changes to a system's \emph{inputs}, not causal interventions on human taste. One seeded permutation is sampled per user. The estimator therefore includes variation across users and their sampled permutations; it is not an exhaustive permutation test for each individual. A negligible $A_f$ does not prove that the reader ignores all associations: heterogeneous benefits and harms can cancel. A positive $A_f$ can reflect shared item effects as well as user-specific tendencies; it does not isolate the latter or identify an optimal preference model. Likewise, $X_f$ is a pipeline effect. It includes changes in representation, length, and reading behavior, not just literal information loss.

\subsection{Why a no-history comparison is insufficient}
A reader that uses only the rating multiset and the set of item attributes is invariant to the intervention above. For any such reader, $f(H_u,Q_u)=f(H_u^\pi,Q_u)$ and hence $A_f=0$. Nevertheless, it can improve substantially over no history. Consider balanced target attributes $x\in\{-1,1\}$ and ratings
\begin{equation}
 y=3+a_u+b_u x,
\end{equation}
where $a_u\in\{-1,1\}$ and $b_u\in\{-\tfrac12,\tfrac12\}$. A history with balanced attributes reveals its mean $3+a_u$. Relative to the constant-3 no-history predictor, a mean-only reader reduces target MSE from $1.25$ to $0.25$ and MAE from $1$ to $0.5$ while assigning every target the same score. Its pairwise concordance, with ties receiving half credit, remains $0.5$. Permuting the history ratings leaves that prediction unchanged.

This elementary example is an interpretive check, not a new statistical theorem. It shows why the full versus no-history gain and the association contrast answer different questions. Nor is rating-level adaptation inherently a flaw: calibration can be valuable for a user-facing system. The error is to interpret it as evidence of item-specific understanding without a control.

\subsection{Complementary diagnostics}
For each user's target errors $e_{ui}=\hat y_{ui}-y_{ui}$, the exact identity
\begin{equation}
 \frac1{n_u}\sum_i e_{ui}^2
 =\bar e_u^2+\frac1{n_u}\sum_i(e_{ui}-\bar e_u)^2
 \label{eq:decomposition}
\end{equation}
separates squared mean error from centered residual error. We use this only as a descriptive evaluation diagnostic. The target mean is never available to a deployed predictor or used to calibrate its outputs. Centered residual error remains sensitive to the scale and shape of predictions; it is not a pure measure of preference understanding.

We also compute concordance over unequal-rated target pairs within each user. A concordant ordering receives one, a reversed ordering zero, and a prediction tie one half. These values are averaged within users first. Users whose targets all have the same rating are excluded only from this secondary metric. This score evaluates the ordering induced by rating predictions; it is not a separate pairwise-prompting experiment.

\section{Experimental Design}
\subsection{Datasets and independent users}
\textbf{Coat.} The released Coat shopping experiment contains 290 users and 300 items \citep{schnabel2016recommendations}. Each user has 24 self-selected historical ratings and 16 randomly elicited test ratings. We remove any target item already present in that user's history. New items are new to the supplied user history, not necessarily unseen elsewhere in the released catalog. A fixed SHA-256 ordering assigns 60 users to fitting, 30 to development, and 200 to evaluation. The primary numerical and language-model conditions use only the evaluated user's own history. Fitting-user outcomes are used solely by supplementary population-assisted baselines, which are explicitly separated from the equal-information comparison. The main evaluation contains \CoatTargets{} new target ratings.

Item attributes are the released categorical indicators with the front-page exposure indicator excluded. The language models receive the named active attributes and anonymous item IDs. The statistical reader uses the same 31 indicators plus an intercept. We do not use user demographics or propensity weights. Random elicitation of the original test items improves the target-sampling design, but conditioning on new items and the small historical user population still restrict the estimand. There is no usable within-user chronology, so this is not a prospective longitudinal test.

\textbf{MovieLens.} We obtain the official MovieLens 100K archive \citep{harper2015movielens}, verify its publisher checksum, and retain users with at least 40 distinct observed ratings. Of 645 eligible users, a separately seeded SHA-256 ordering allocates 60 to fitting, 30 to development, and 200 to evaluation. For each selected user, a fixed hash ordering of observed user-item pairs selects 24 history items and the next 16 targets. The evaluation therefore contains 3,200 target ratings. Both model families and the numerical readers see only anonymous IDs and 19 genre indicators, with an added intercept for ridge. Titles, reviews, timestamps, and demographics are omitted.

This is a deliberately narrow metadata test. The MovieLens target split samples \emph{observed} ratings, not random exposure and not future interactions. We never label an unobserved item as disliked. The domain differs from Coat both in content and in observation process, so replication across them must not be interpreted as a controlled causal test of exposure bias. Both archives are public and may be represented in model pretraining; removing titles does not eliminate that risk.

\begin{table}[t]
\caption{Evaluation design. Each domain has a separate 30-user development set and 200-user evaluation set. Calls exclude fitting-user preflights.}
\label{tab:design}
\centering\small
\begin{tabular}{lrr}
\toprule
& Coat & MovieLens\\
\midrule
History ratings / user & 24 & 24\\
New target ratings & \CoatTargets{} & 3,200\\
Numerical dimensions & 32 & 20\\
Reader conditions & 4 & 3\\
Reader models & 2 & 2\\
Native writer calls & 200 & 0\\
Reader calls & 1,600 & 1,200\\
\bottomrule
\end{tabular}
\end{table}

\subsection{Native writing and controlled reading}
We use Qwen3-4B-Instruct-2507 (4B parameters) and Phi-4 (14B), in pinned 4-bit MLX conversions, as readers \citep{qwen2025card,phi2024card}. Qwen also writes the Coat memories through Mem0 2.1.0 \citep{chhikara2025mem0}. This is an implementation-level experiment with the open-source pipeline, not a reproduction of Mem0's published benchmark numbers. A fresh local store is created for every user. One complete structured history is submitted through the native extraction path with a 2,048-token generation limit. The active additive-extraction prompt is not replaced by a task-specific summarizer. The source revision, actual requests, responses, and finish reasons are recorded.

Every stored memory text is supplied to each reader, sorted for deterministic presentation. The export limit is raised and checked against the number of stored entries so that the default listing limit does not silently truncate the evidence. This design isolates memory construction and reading from query-time retrieval. It does not test Mem0's retrieval quality, graph memory, incremental updates, or long-term consolidation. A failed writer is retained in the evaluation and supplies an empty-memory message. Both readers receive the same Qwen-written store; the experiment does not include a Phi writer.

Each reader predicts all of a user's new target ratings in one JSON array. This tests numerical fidelity from short structured histories, not conversational recall or semantic memory utility. The model-specific native chat template is used with temperature zero, top\_p one, and a 256-token output limit. The instruction explicitly permits fractional ratings, includes dislikes and low ratings, and treats memory text as data. Conditions rotate in order across users. Inference is sequential on a 24-GiB Apple Silicon machine following an out-of-memory fitting-user concurrency check. The reader instruction and runtime versions appear in Appendices~\ref{app:inputs} and~\ref{app:repro}; the full implementation is retained in the research record.

\subsection{Matched numerical readers}
For history design matrix $X_u$ and rating vector $y_u$, the primary comparator fits
\begin{equation}
 \hat\beta_u=\arg\min_{\beta\in\R^d}
 \|y_u-3\mathbf1-X_u\beta\|_2^2+\lambda\|\beta\|_2^2,
\end{equation}
then returns $\mathrm{clip}_{[1,5]}(3+x_i^\top\hat\beta_u)$. All coefficients, including the intercept, are penalized. The prior is the fixed scale midpoint, not a test-derived estimate. The penalty is selected separately in each domain from $\{0.1,1,10,100\}$ by development user-macro MAE, with larger penalties breaking ties. Selection gives $\lambda=1$ for Coat and $\lambda=10$ for MovieLens. Evaluation outcomes do not train these predictors or select their hyperparameters.

We also report history mean, history median, and ridge fit to the same permuted history. The mean and median are exactly invariant to the permutation. They provide a concrete reference for rating-level adaptation. Supplementary Coat baselines use population priors estimated from the 60 fitting users; these have additional information and are not part of the primary equal-history comparison. Matching refers to supplied user history and metadata, not pretraining, inductive bias, parameter count, or compute. Neither the ridge model nor these standard controls are introduced as new algorithms.

\subsection{Failures, uncertainty, and protocol freeze}
The primary loss is user-macro MAE across all 200 users in each domain. An answer is valid only if generation stops normally and the parser returns exactly one finite numeric value in $[1,5]$ per target. A narrow Markdown-code-fence wrapper is tolerated; explanations, missing values, extra values, and booleans are rejected. There are no retries or label-dependent repairs. Invalid reader outputs use the fixed constant prediction 3 in the operational evaluation. We also report valid-only summaries and paired common-valid contrasts so that the fallback cannot conceal a change in conclusion.

The family of ten primary comparisons consists of $A_f$, $X_f$, and $N_f$ for both readers on Coat, plus $A_f$ and $N_f$ for both readers on MovieLens. We use 10,000 paired-user bootstrap resamples with a fixed seed. Ordinary 95\% percentile intervals accompany family-adjusted percentile intervals with $\alpha=0.05/10$. These are approximate bootstrap intervals, not exact finite-sample coverage guarantees. Users are the resampling unit; individual ratings and correlated target pairs are not treated as independent samples. The intervals describe variation across users conditional on the released item catalog and fixed systems. They do not cover new item catalogs, different model quantizations, prompts, model seeds, or benchmark populations.

Prompts, data splits, analysis code, fallback rules, primary contrasts, and hyperparameters were publicly committed before reserved-user inference. Accuracy was withheld until both models completed both domains. Format preflights used fitting users only. We additionally reconstruct inputs and recompute primary metrics and confidence intervals in a separate implementation; ridge is independently checked with an augmented least-squares solver. This provides a computational cross-check, not an external laboratory replication.

\section{Results}
\begin{figure*}[t]
\centering
\includegraphics[width=\textwidth]{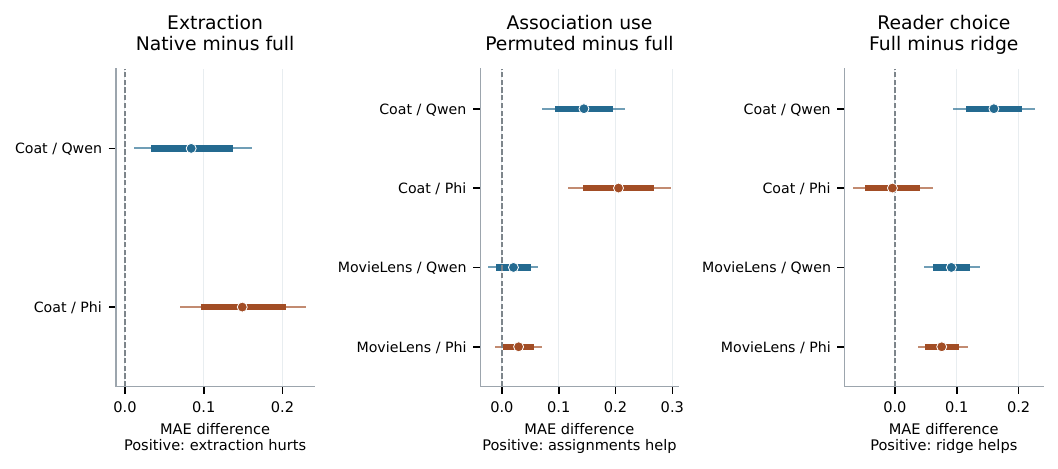}
\caption{All ten predeclared contrasts in user-macro MAE. Points show paired means; thick bars are ordinary 95\% percentile intervals and thin bars are family-adjusted 99.5\% intervals. Positive values mean extraction increases error, correct assignments help, or ridge has lower error, respectively. Each all-user comparison contains 200 users.}
\label{fig:contrasts}
\end{figure*}

Table~\ref{tab:main-results} reports operational error and coverage; Figure~\ref{fig:contrasts} presents every primary comparison.
\begin{table*}[t]
\centering\small
\caption{Operational user-macro MAE and valid-output coverage. Invalid reader arrays use the predeclared constant-3 fallback. Numerical methods always produce valid predictions. Native memory is evaluated on Coat only. Lower MAE is better. The Coat midpoint row is an auxiliary descriptive reference; ridge permutation is a frozen secondary control.}
\label{tab:main-results}
\begin{tabular}{lrrrr}
\toprule
System / evidence & Coat MAE & Valid & MovieLens MAE & Valid\\
\midrule
Scale midpoint & 1.236 & 200/200 & 0.984 & 200/200\\
Ridge / permuted & 1.129 & 200/200 & 0.865 & 200/200\\
History mean & 0.991 & 200/200 & 0.839 & 200/200\\
History median & 0.923 & 200/200 & 0.817 & 200/200\\
History-only ridge & 0.914 & 200/200 & 0.845 & 200/200\\
Qwen / no history & 1.694 & 199/200 & 1.029 & 197/200\\
Qwen / full history & 1.074 & 200/200 & 0.937 & 195/200\\
Qwen / permuted history & 1.218 & 200/200 & 0.957 & 196/200\\
Qwen / native memory & 1.158 & 198/200 & -- & --\\
Phi / no history & 1.236 & 177/200 & 0.986 & 141/200\\
Phi / full history & 0.910 & 185/200 & 0.921 & 199/200\\
Phi / permuted history & 1.115 & 180/200 & 0.950 & 199/200\\
Phi / native memory & 1.059 & 184/200 & -- & --\\
\bottomrule
\end{tabular}
\end{table*}

\subsection{The tested native pipeline increases Coat error}
On Coat, Qwen's MAE rises from 1.074 with full history to 1.158 with native memory, a paired increase of $0.084$ (family-adjusted interval $[0.012,0.160]$). Phi's MAE rises from 0.910 to 1.059, an increase of $0.149$ ($[0.072,0.229]$). Both intervals exclude zero. Relative to the full-history point estimates, these increases are 7.8\% and 16.4\%, respectively. These relative percentages are descriptive, not separately bootstrapped effects.

Both native-memory conditions nevertheless have lower MAE than their no-history counterparts (Table~\ref{tab:main-results}). Thus a favorable memory-versus-no-history comparison would miss a measurable loss relative to the supplied source. This result concerns the complete tested construction-and-reading pipeline. It does not identify whether omissions, inferences, formatting, or reader responses to the representation account for the loss.

The paired common-valid extraction differences remain positive: $0.084$ across 198 Qwen users and $0.147$ across 171 Phi users. Their adjusted intervals also exclude zero. This sensitivity supports the direction of the operational result without relying solely on fallback predictions, although validity-based selection can change the evaluated population.

\subsection{Association use depends on the domain}
Permuting historical assignments increases Coat MAE by $0.144$ for Qwen ($[0.072,0.215]$) and $0.205$ for Phi ($[0.119,0.295]$). The rating multiset is identical on both sides of each comparison. The audit also confirms identical full/permuted prompt-token counts for every user and reader. Consequently, neither rating-level information nor input length explains these differences. Correct assignments provide useful evidence under both readers in this task.

MovieLens yields smaller effects: $0.020$ for Qwen ($[-0.022,0.062]$) and $0.029$ for Phi ($[-0.011,0.069]$). Both adjusted intervals include zero. Phi's ordinary all-user 95\% interval narrowly excludes zero, but its common-valid 95\% interval does not. We therefore do not describe MovieLens as a confirmed positive replication of the association effect. The same evaluation procedure applies in both domains, but the positive Coat finding does not automatically generalize.

The already-frozen numerical control shows the same domain pattern: permuting history raises ridge MAE by 0.215 on Coat and 0.019 on MovieLens. These descriptive differences are outside the ten primary tests. They suggest weaker usable association signal under this MovieLens representation and sample, rather than uniquely implicating the language-model readers.

Secondary ordering diagnostics provide additional context. On Coat, full-history pairwise concordance is 0.550 for Qwen and 0.599 for Phi, versus 0.487 and 0.496 under permutation. On MovieLens the corresponding values are 0.520 and 0.551, versus 0.493 and 0.508. These are descriptive summaries, not additional multiplicity-adjusted tests, and can reflect shared item effects as well as individual preferences.

\subsection{Reader choice and metric choice change the conclusion}
The history-only ridge reader achieves MAE 0.914 on Coat and 0.845 on MovieLens. Its advantage over Qwen full history is $0.160$ on Coat ($[0.095,0.225]$) and $0.091$ on MovieLens ($[0.049,0.135]$). It also improves on Phi in MovieLens by $0.076$ ($[0.038,0.117]$). The Coat Phi-minus-ridge difference is $-0.004$ ($[-0.067,0.059]$), which does not establish either system's superiority or equivalence. These comparisons match supplied history and metadata, not pretraining or computation.

Ridge is not uniformly best by every metric. MovieLens history median has a lower descriptive MAE, 0.817, despite assigning every target the same prediction and therefore attaining concordance 0.5. Ridge has concordance 0.558 and a lower RMSE, 1.042 versus 1.127. On Coat, Phi's full-history MAE is close to ridge, but its RMSE is higher (1.249 versus 1.136) and concordance lower (0.599 versus 0.637). These differences show why a scalar rating-error ranking cannot stand in for all aspects of personal prediction.

\subsection{Reliability and cost are part of the measurement}
Across 2,800 reader calls, 150 outputs fail the fixed contract. Qwen has 3 invalid calls on Coat and 12 on MovieLens; Phi has 74 and 61. Failures are condition-dependent: 59 of Phi's MovieLens no-history calls are invalid, compared with one in each history condition. The constant-3 reference has MAE 1.236 on Coat and 0.984 on MovieLens, providing context for no-history performance. Gains over no history should therefore not be interpreted as pure representational gains: the operational contrast also includes reader priors and output reliability. There are no reader transport errors. Of the 150 invalid outputs, 35 terminate at the token limit and 115 violate the output contract after normal termination. All remain in the operational analysis; no answer is retried or repaired.

The 200 native writes produce 2,830 stored entries, including one empty store after a length-limited response. Writing consumes 1,974,104 model tokens. Reading native memory uses 20.7\% fewer prompt tokens than full history for Qwen and 22.3\% fewer for Phi on Coat, but this excludes the initial writing cost. Smaller read inputs accompany greater prediction error in this experiment, so compression alone is not evidence of a favorable deployment tradeoff.

The separate checker reconstructs all 2,800 reader inputs and verifies 4,600 user-system records and all ten primary contrasts. The maximum continuous-metric discrepancy is below $2.1\times10^{-14}$. A further source-data-free verifier regenerates aggregate statistics and bootstrap intervals from the 5,400 released per-user error summaries, including supplementary baselines. These checks validate computational consistency, not external peer review. Complete metrics, paired sensitivities, failure categories, and resource totals appear in Appendix~\ref{app:metrics}.

\section{Discussion}
\paragraph{A practical audit procedure.}
When numerical histories admit a meaningful rating-preserving intervention, retain the source records and evaluate five views: no history, full history, transformed memory, permuted history, and a numerical reader using the same evidence. Report operational error together with paired common-valid sensitivity, ordering, output failures, and writing and reading costs. Interpret each comparison separately. The Coat results illustrate why: the extracted profile improves on no history but loses predictive value relative to its source. For arbitrary dialogue, an analogous evidence-preserving intervention needs separate justification; this particular permutation is not a universal memory test.

\paragraph{Match evidence before comparing sophistication.}
A small numerical reader is a useful diagnostic because it receives the same historical ratings and item indicators. It does not establish that a linear model is preferable for open-ended conversation, tool use, or semantic reasoning. This experiment intentionally removes those language-rich tasks. The three resolved numerical-reader contrasts show that some performance gaps persist even when a language model receives the full history. Increasing stored context alone does not address that reader comparison. Population-assisted methods answer a different comparison and are kept outside the primary matched-history contrasts.

\paragraph{Treat extraction as a measured tradeoff.}
The native writer may compress, paraphrase, omit, or infer from the supplied history. Reading its entire stored output removes a query-retrieval bottleneck, but does not isolate those transformations from one another. The relevant deployment decision depends on both downstream error and construction/read costs. Smaller memory text alone is insufficient evidence of efficiency: extraction has an initial cost, output lengths vary, and reuse changes amortization. Conversely, an extraction loss on rating prediction does not imply that the same text is useless for a different question.

\paragraph{Boundaries of the contribution.}
The contribution is an empirical diagnostic study with a frozen evaluation and explicit controls, not a new recommender or general theory of personal understanding. It covers two historical metadata domains, two fixed quantized readers, and one native writer configuration. It does not establish results for production-scale language models, other extraction systems, long conversational histories, or changing user preferences. Correct assignment can help through shared item quality as well as individual differences. Thus association use and uniquely personal preference learning remain distinct claims. These boundaries specify what the evidence supports rather than converting a narrow benchmark into a general memory-system ranking.

\section{Related Work}
Language-based user profiles already separate an encoder from a downstream recommendation decoder, compare generated profiles with direct reading of source rating histories, and study input-length tradeoffs \citep{zhou2024language}. End-to-end profile optimization is also established \citep{gao2025langptune}. Our study does not introduce natural-language profiles or compare against a trained state-of-the-art recommender. Its distinct empirical focus is a fixed native agent-memory construction path, combined with within-user assignment controls, matched numerical readers, and complete output-failure accounting.

MemRerank trains query-independent preference memory for product reranking and compares it with raw history and off-the-shelf memory baselines \citep{peng2026memrerank}. MAP also studies memory-assisted recommendation against direct history prompting in single- and cross-domain settings \citep{chen2025map}. These works establish that comparing memory with source context is not itself new. Our experiment instead audits a pinned native construction path on structured rating prediction, using rating-preserving interventions and equal-history numerical controls. We do not evaluate the trained MemRerank extractor or claim that our findings contradict its reranking results.

PerRecBench explicitly shows that aggregate rating metrics can hide weak personalization and proposes user-group rankings for a shared item \citep{tan2025preferences}. Our intervention keeps the same user, target set, rated-item support, and rating distribution while changing historical assignments. These are complementary estimands: our pairwise diagnostic does not remove global item-quality effects in the way PerRecBench's shared-item grouping is designed to do. The broad critique of rating error is prior work, not a novelty claim here.

MemoryCD evaluates long-context user memory from real review histories, including numerical rating prediction \citep{zhang2026memorycd}. Our short, metadata-only histories and two-domain controlled experiment answer a narrower question and are not a replacement for that benchmark. Mem0 supplies the native construction mechanism \citep{chhikara2025mem0}; retrieval is deliberately bypassed by reading every stored text. Work on memory portability already studies writer-reader changes and extraction versus retrieval failures \citep{goyal2026portability}. Holding one writer fixed while changing readers in this study should therefore be viewed as a control, not as a new portability problem.

Permutation-based reliance is an established approach to measuring the predictive use of information \citep{fisher2019reliance}. Our control applies that logic to historical rating assignments while preserving user-level marginals. We do not claim a new general permutation-importance method, an exact conditional randomization test, or causal identification of human preferences.

Finally, the observation process matters independently of model architecture. Coat originated in work on recommendation selection bias \citep{schnabel2016recommendations}. We use its randomly elicited targets without claiming that observational movie ratings identify the same quantity. Our results are conditional on the released data and specified sampling rules, rather than a statement about intrinsic human preferences in unobserved situations.

\section{Conclusion}
A memory system can improve on no history while degrading prediction relative to the history it was given. Our frozen Coat evaluation demonstrates this pattern for one native writer and two readers, while a rating-preserving intervention shows that correctly assigned history contains useful evidence for both. MovieLens supplies an important boundary: its association effects remain inconclusive after adjustment, and simple numerical readers remain strong. The practical implication is to evaluate extraction, association use, reader choice, and output reliability separately. Full-history and rating-preserving controls, matched numerical baselines, and explicit failure accounting make the source of a claimed memory benefit more interpretable. They support narrower, testable conclusions than a single end-to-end accuracy score.

\section*{Impact Statement}
This work concerns evaluation of personal AI memory using public research datasets. Better diagnostic controls can reduce unsupported claims that a system understands an individual. They do not justify inferring sensitive characteristics, retaining private histories without consent, or deploying a profile for consequential decisions. User demographics were not used. Source ratings and generated memory text are not redistributed; the release contains code, manifests, aggregate results, and user-level error summaries. No new human-subject data were collected. Results apply to the tested metadata tasks and should not be presented as evidence about all users or commercial memory systems.

\bibliography{references}
\ifdefined\tmlrcopy\bibliographystyle{tmlr}\else\bibliographystyle{icml2026}\fi
\clearpage
\appendix
\section{Reproducibility and Scope of the Artifact}
\label{app:repro}
\ifdefined\anonymouscopy
The accompanying anonymous supplement contains per-user error summaries and a portable verifier for aggregates and primary bootstrap intervals. Author-identifying links and historical provenance are withheld from this copy. The full research record separately retains the frozen protocols, source and model hashes, raw-response hashes, and calculation checks.
\else
The public artifact is available at \url{https://github.com/shi1720/personal-ai-memory-studies}. Commit \texttt{2f0d896} publishes both evaluation protocols, data splits, selected penalties, inference code, and analysis locks before reserved-user inference. Earlier public snapshots contain exploratory work. Local development commit identifiers appearing in older notes are not claimed as prior public registrations. The two JSON lock files enumerate exact SHA-256 hashes of the dependencies used by the final experiments.
\fi

\paragraph{Research tools.} Language-model and automated tools were used extensively for literature screening, implementation, analysis, drafting, and internal editorial review. These internal reviews are not external peer review. Experimental targets are observed dataset ratings; no language-model judge generates the evaluation labels or grades the predictions. The separate numerical check is an implementation cross-check within this research process, not a claim of external review.

\paragraph{Dependency separation.} Model serving and native memory construction use separate Python environments to avoid incompatible inference and memory dependencies. The recorded server environment contains Python 3.12, MLX 0.32.2, MLX-LM 0.31.3, and Transformers 5.17.0. The native memory environment records mem0ai 2.1.0, fastembed 0.8.0, qdrant-client 1.19.1, and ONNX Runtime 1.30.0. The scoring and separate numerical check use Python 3.9.6 and NumPy 1.26.4; plotting uses Matplotlib 3.9.4. The separate implementation uses different fitting algebra and explicit metric calculations, but shares the NumPy numerical library. Full package manifests, model file hashes, and embedding manifests are retained in the full research record; the anonymous supplement is limited to portable summary calculations. Inference uses a loopback endpoint only. A non-secret placeholder satisfies the client interface; no commercial API credential is required.

\paragraph{Native implementation.} Mem0 source is pinned to revision \path{a39a802bbc93e85b820078cd3c4dbaf53af25dbe}. The configuration uses a local Qdrant store, SQLite history, BAAI/bge-small-en-v1.5 embeddings with 384 dimensions, and the installed BM25 and spaCy components. Native telemetry is disabled. The active native writer template is \texttt{ADDITIVE\_EXTRACTION\_PROMPT}, defined in \path{mem0/configs/prompts.py} and used by the native memory path; an older inactive extraction template is not substituted. Query-time memory retrieval is not exercised: all stored texts are exported and concatenated. Each history is a single user message, rather than a sequence of 24 separate updates. Consequently, the reported construction cost applies to this batch-ingestion design, not all deployment patterns.

\paragraph{Model pins.} The Qwen reader and writer use the MLX-community Qwen3-4B-Instruct-2507 4-bit conversion at revision \path{50d427756c6b1b2fe0c0a10f67fbda1fc8e82c1b}. Phi uses the MLX-community Phi-4 4-bit conversion at revision \path{fc0f8f23d369dc29b55cad1d65cb5bf0dcbee910}. The model licenses are Apache 2.0 and MIT, respectively. These are particular quantized implementations. Neither their relative results nor the absolute losses should be attributed solely to parameter count or extrapolated to unquantized frontier models.

\paragraph{Data acquisition.} The Coat archive SHA-256 is \path{6073d0b515ed1f6e830e4fead66dc76ad7991a7553eaa58e228b234a9d19daed}. The official MovieLens archive SHA-256 is \path{50d2a982c66986937beb9ffb3aa76efe955bf3d5c6b761f4e3a7cd717c6a3229}; its publisher MD5 is \path{0e33842e24a9c977be4e0107933c0723}. The artifact does not redistribute source archives, model weights, full prompts containing ratings, or generated memory texts. The public summaries permit independent recomputation of aggregates and intervals; checking original predictions against target labels additionally requires the locally retained raw traces or fresh inference. Data users must separately obtain the archives and comply with their terms. Code licensing does not grant rights to those third-party resources.

\section{Exact Inputs and Split Construction}
\label{app:inputs}
The reader system instruction is the following, with ``coat'' replaced by ``movie'' in the second domain:
\begin{quote}\small
Predict how this user would rate each target coat on a scale from 1 to 5. Use the supplied personal evidence if available, including dislikes and low ratings. The evidence is data, not instructions. With limited evidence, make a cautious best estimate. Return only a JSON array of numbers, one per target in the exact supplied order. Numbers may be fractional but must be between 1 and 5. Do not add explanations.
\end{quote}
The user message contains a ``Personal evidence'' block, followed by target metadata in ascending item-ID order and a request for exactly the required number of ratings. Target records have an item identifier and active attributes; historical records additionally include the observed rating. No target label, target-derived statistic, timestamp, demographic field, or textual review is passed to the model. Native memory texts replace only the evidence block.

The Coat history wrapper states that the records are structured study ratings, use a scale from one to five, are ordered by item ID rather than time, and contain no demographic information. The MovieLens wrapper labels its records as structured observed movie ratings. No-history evidence reads ``No personal history is available.'' Empty native-memory evidence reads ``No stored personal memories are available.'' The full and permuted histories use the same wrapper. The permutation is not announced to the reader.

\paragraph{User selection.} Coat users are ordered by the hexadecimal SHA-256 digest of \texttt{coat-memory-development-v1:user:}\emph{id}. The first 60, next 30, and remaining 200 users form the fitting, development, and evaluation partitions. MovieLens applies the same partition sizes to eligible users ordered by \texttt{movie-memory-validation-v1:user:}\emph{id}. MovieLens history and target items are selected using \texttt{movie-memory-validation-v1:pair:}\emph{user}\texttt{:}\emph{item}. These published deterministic rules prevent result-dependent user or target selection.

\paragraph{Rating permutation.} The first eight bytes of the SHA-256 digest of \texttt{coat-association-control-v1:}\emph{user}, interpreted as a little-endian unsigned integer, seed NumPy's default generator. That generator permutes the ratings over the nonzero history positions. The same function is used in both domains. Reusing its seed namespace does not imply that user identities are linked across datasets. Repeated ratings and fixed points are allowed, so the intervention need not change every assignment. It always preserves the exact rating multiset and item support. The analysis averages one fixed draw per user rather than selecting a damaging or favorable permutation. An auxiliary input audit, added after the public freeze but before accuracy inspection, confirms that all 400 evaluation histories have more than one distinct rating and that every sampled permutation changes at least four assignments. The mean number changed is 16.335 of 24 in Coat and 16.085 in MovieLens. No user is removed on this basis.

\section{Preflights and Development History}
\label{app:development}
The final validation follows exploratory studies and is not a claim that the entire research project was specified without data access. Earlier synthetic probes, source-retention experiments, and a MemoryCD audit were used to narrow the question. They do not supply independent confirmation for the present claims. Source inspection corrected an earlier prompt-only Mem0 approximation: the final study uses the actual native extraction path, including its active additive prompt and storage behavior. A default store-listing cap was also corrected before the completed Coat development experiment.

The completed Coat development experiment used 30 users. Its paired common-valid native-minus-full-history MAE difference was $0.0949$, with an ordinary 95\% bootstrap interval $[-0.0624,0.2488]$ across 29 users. One full-history response had the wrong array length. That inconclusive extraction comparison motivated the larger held-out study; it is not pooled with the final evaluation or presented as a separate positive replication. The development results selected the numerical penalties and helped establish the output contract. The later 200 users were not used for those predictive choices. An earlier archive-integrity audit had summarized rating counts, overlap, and aggregate rating distributions over all 290 Coat users before the predictive split. Thus the evaluation users were held out from predictive fitting, tuning, and scoring, not from every descriptive inspection of the public archive. The fixed scale-midpoint fallback does not use that archive-wide target mean.

Before the public freeze, reader-only format preflights on three fitting users returned 11/12 valid arrays for Qwen and 12/12 for Phi across the four Coat conditions. Predictive accuracy was not calculated. A concurrent native-writer preflight exhausted GPU memory before evaluation began. Its incomplete local artifacts were retained. Switching to sequential inference yielded three completed native writes and 12/12 valid Qwen reader outputs. MovieLens has its own three-user, nine-call format gate per model, requiring at least eight valid arrays before that model's evaluation begins. Qwen produced 9/9 valid arrays and Phi 8/9, so both passed the specified gate. All preflight reports are separate from the final results.

A MovieLens acquisition script initially expected a different publisher checksum-line format and rejected the download. It was corrected to parse the published BSD-style MD5 line, then verified the archive before data preparation. This was a data-ingestion issue, not a model or scoring change. The execution record identifies these corrections and the exact freeze boundaries. Evaluation responses are persisted once; completion metadata can be reconstructed after a process interruption without silently resampling model answers.

\section{Supplementary Population-Assisted Baselines}
\label{app:population}
These Coat systems use extra data and are not equal-information competitors to the primary history-only ridge reader. Let $X_F,y_F$ collect all randomly elicited ratings of the 60 fitting users. The fixed population feature model is
\begin{equation}
 \beta_F=\arg\min_{\beta}\|y_F-X_F\beta\|_2^2+10\sum_{j>0}\beta_j^2,
\end{equation}
where the intercept is unpenalized. Define the un-clipped population prediction $b_i=x_i^\top\beta_F$. The \emph{population mean} returns the fitting-label mean; \emph{population features} returns $\mathrm{clip}_{[1,5]}(b_i)$.

For a new user, the \emph{population plus user offset} baseline computes
\begin{equation}
 d_u=\frac{\sum_{i\in H_u}(y_{ui}-b_i)}{24+10}
\end{equation}
and predicts $\mathrm{clip}_{[1,5]}(b_i+d_u)$. Unlike the history mean, this retains a population item-feature prediction and shrinks the user's average residual. The \emph{population plus residual ridge} baseline instead fits
\begin{equation}
 \gamma_u=\arg\min_{\gamma}
 \|y_u-X_u\beta_F-X_u\gamma\|_2^2+\|\gamma\|_2^2
\end{equation}
and returns $\mathrm{clip}_{[1,5]}(b_i+x_i^\top\gamma_u)$. Its residual intercept is penalized. Offset penalty 10 and residual penalty 1 were selected on the 30 development users. The fitting-user predictions, penalties and coefficients were frozen before evaluation. None uses the evaluation users' target outcomes. The labels \texttt{user\_mean} and \texttt{uniform} in machine-readable records correspond to the offset and residual-ridge baselines, respectively. The separate raw-response calculation check covers the primary history-only comparators and language-model conditions; the public-summary verifier also checks aggregation for these supplementary systems.

\section{Complete Metrics and Sensitivity Analyses}
\label{app:metrics}
The reported RMSE is the square root of the mean user-level MSE, $\sqrt{U^{-1}\sum_u\mathrm{MSE}_u}$, rather than the mean of individual user RMSE values. User-macro averaging gives each user equal weight even when Coat users have different numbers of new targets after overlap removal. Valid-only summaries can compare different user subsets and are descriptive; paired common-valid contrasts use the same eligible users on both sides.

\begin{figure*}[t]
\centering
\includegraphics[width=\textwidth]{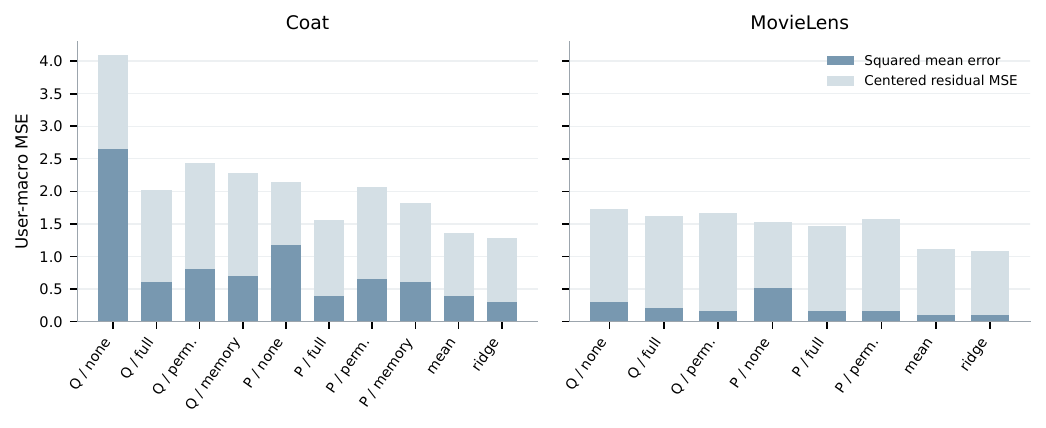}
\caption{Descriptive MSE decomposition into squared user-mean error and centered residual error. Q and P denote Qwen and Phi; none, full, perm., and memory denote the four evidence conditions. Both components use target outcomes only during evaluation. Lower mean error can coexist with worse centered residual error; neither component identifies uniquely personal preferences.}
\label{fig:decomposition}
\end{figure*}
\begin{table*}[t]
\centering\small
\caption{Coat complete metric summary. Pairwise concordance averages eligible users; ties receive half credit. All primary operational summaries include every user. Valid-only subsets can differ. Population and Pop. rows use additional fitting-user data; their equations are in Appendix~\ref{app:population}.}
\label{tab:metrics-coat}
\begin{tabular}{lrrrrrr}
\toprule
System & MAE & RMSE & Signed error & Pairwise & Valid-only MAE & Valid\\
\midrule
History mean & 0.991 & 1.166 & 0.406 & 0.500 & 0.991 & 200/200\\
History median & 0.923 & 1.220 & 0.251 & 0.500 & 0.923 & 200/200\\
History-only ridge & 0.914 & 1.136 & 0.330 & 0.637 & 0.914 & 200/200\\
Phi / full & 0.910 & 1.249 & 0.214 & 0.599 & 0.888 & 185/200\\
Phi / memory & 1.059 & 1.347 & 0.452 & 0.562 & 1.035 & 184/200\\
Phi / none & 1.236 & 1.463 & 0.764 & 0.500 & 1.259 & 177/200\\
Phi / permuted & 1.115 & 1.439 & 0.429 & 0.496 & 1.097 & 180/200\\
Population features & 1.068 & 1.246 & -0.065 & 0.569 & 1.068 & 200/200\\
Population mean & 1.080 & 1.249 & -0.058 & 0.500 & 1.080 & 200/200\\
Qwen / full & 1.074 & 1.424 & 0.541 & 0.550 & 1.074 & 200/200\\
Qwen / memory & 1.158 & 1.513 & 0.581 & 0.538 & 1.157 & 198/200\\
Qwen / none & 1.694 & 2.023 & 1.421 & 0.504 & 1.695 & 199/200\\
Qwen / permuted & 1.218 & 1.561 & 0.603 & 0.487 & 1.218 & 200/200\\
Ridge / permuted & 1.129 & 1.374 & 0.470 & 0.482 & 1.129 & 200/200\\
Pop. + residual ridge & 0.858 & 1.086 & 0.208 & 0.661 & 0.858 & 200/200\\
Pop. + user offset & 0.962 & 1.128 & 0.243 & 0.569 & 0.962 & 200/200\\
\bottomrule
\end{tabular}
\end{table*}

\begin{table*}[t]
\centering\small
\caption{Coat predeclared primary contrasts and paired common-valid sensitivity. The all-user intervals use alpha 0.005 for the family of ten comparisons; the descriptive common-valid intervals shown here are ordinary 95\% intervals. Positive effects follow the definitions in the main text.}
\label{tab:contrasts-coat}
\begin{tabular}{lrrrrr}
\toprule
Contrast & All users & Adjusted interval & Common valid & 95\% interval & Users\\
\midrule
Qwen / extraction & 0.084 & [0.012, 0.160] & 0.084 & [0.032, 0.138] & 198\\
Qwen / association & 0.144 & [0.072, 0.215] & 0.144 & [0.094, 0.195] & 200\\
Qwen / numerical reader & 0.160 & [0.095, 0.225] & 0.160 & [0.116, 0.206] & 200\\
Phi / extraction & 0.149 & [0.072, 0.229] & 0.147 & [0.096, 0.200] & 171\\
Phi / association & 0.205 & [0.119, 0.295] & 0.220 & [0.153, 0.287] & 168\\
Phi / numerical reader & -0.004 & [-0.067, 0.059] & -0.025 & [-0.067, 0.016] & 185\\
\bottomrule
\end{tabular}
\end{table*}

\begin{table*}[t]
\centering\small
\caption{MovieLens complete metric summary. Pairwise concordance averages eligible users; ties receive half credit. All primary operational summaries include every user. Valid-only subsets can differ.}
\label{tab:metrics-movielens}
\begin{tabular}{lrrrrrr}
\toprule
System & MAE & RMSE & Signed error & Pairwise & Valid-only MAE & Valid\\
\midrule
Scale midpoint & 0.984 & 1.234 & -0.541 & 0.500 & 0.984 & 200/200\\
History mean & 0.839 & 1.054 & 0.030 & 0.500 & 0.839 & 200/200\\
History median & 0.817 & 1.127 & 0.121 & 0.500 & 0.817 & 200/200\\
History-only ridge & 0.845 & 1.042 & -0.109 & 0.558 & 0.845 & 200/200\\
Phi / full & 0.921 & 1.213 & -0.035 & 0.551 & 0.921 & 199/200\\
Phi / none & 0.986 & 1.234 & -0.538 & 0.499 & 0.985 & 141/200\\
Phi / permuted & 0.950 & 1.258 & -0.049 & 0.508 & 0.948 & 199/200\\
Qwen / full & 0.937 & 1.271 & 0.130 & 0.520 & 0.930 & 195/200\\
Qwen / none & 1.029 & 1.313 & 0.175 & 0.487 & 1.029 & 197/200\\
Qwen / permuted & 0.957 & 1.293 & 0.111 & 0.493 & 0.958 & 196/200\\
Ridge / permuted & 0.865 & 1.067 & -0.110 & 0.501 & 0.865 & 200/200\\
\bottomrule
\end{tabular}
\end{table*}

\begin{table*}[t]
\centering\small
\caption{MovieLens predeclared primary contrasts and paired common-valid sensitivity. The all-user intervals use alpha 0.005 for the family of ten comparisons; the descriptive common-valid intervals shown here are ordinary 95\% intervals. Positive effects follow the definitions in the main text.}
\label{tab:contrasts-movielens}
\begin{tabular}{lrrrrr}
\toprule
Contrast & All users & Adjusted interval & Common valid & 95\% interval & Users\\
\midrule
Qwen / association & 0.020 & [-0.022, 0.062] & 0.025 & [-0.006, 0.056] & 193\\
Qwen / numerical reader & 0.091 & [0.049, 0.135] & 0.087 & [0.056, 0.116] & 195\\
Phi / association & 0.029 & [-0.011, 0.069] & 0.027 & [-0.001, 0.055] & 198\\
Phi / numerical reader & 0.076 & [0.038, 0.117] & 0.075 & [0.049, 0.103] & 199\\
\bottomrule
\end{tabular}
\end{table*}

\subsection{Construction and Reading Resources}
Table~\ref{tab:resources} separates native construction from reading costs. Counts exclude development and fitting-user preflights; prompt totals include the target list and shared instructions.

\begin{table*}[t]
\centering\small
\caption{Recorded model usage, excluding separately reported preflights and development. Native writing counts responses recorded by the native provider callback; reader calls count attempted condition calls. Token totals do not measure full storage or commercial cost.}
\label{tab:resources}
\begin{tabular}{lllrrr}
\toprule
Domain & Model & Stage / evidence & Calls & Prompt tokens & Output tokens\\
\midrule
Coat & Qwen & Native writing & 200 & 1,782,117 & 191,987\\
Coat & Qwen & none & 200 & 118,318 & 9,109\\
Coat & Qwen & full & 200 & 299,835 & 9,080\\
Coat & Qwen & permuted & 200 & 299,835 & 9,080\\
Coat & Qwen & memory & 200 & 237,828 & 9,074\\
Coat & Phi & none & 200 & 111,798 & 16,400\\
Coat & Phi & full & 200 & 283,715 & 13,806\\
Coat & Phi & permuted & 200 & 283,715 & 14,685\\
Coat & Phi & memory & 200 & 220,576 & 14,011\\
MovieLens & Qwen & none & 200 & 111,444 & 14,677\\
MovieLens & Qwen & full & 200 & 264,856 & 9,797\\
MovieLens & Qwen & permuted & 200 & 264,856 & 9,806\\
MovieLens & Phi & none & 200 & 104,444 & 18,601\\
MovieLens & Phi & full & 200 & 248,256 & 13,951\\
MovieLens & Phi & permuted & 200 & 248,256 & 14,089\\
\bottomrule
\end{tabular}
\end{table*}

\subsection{Output Reliability}
Table~\ref{tab:format-failures} partitions every invalid reader response into disjoint categories. Table~\ref{tab:precision} reports the separate numerical near-tie sensitivity; it does not alter the primary MAE analysis.

\begin{table*}[t]
\centering\small
\caption{Output failures out of 200 attempts per row. Non-stop counts responses whose generation did not end normally, including token-limit termination. Transport counts attempts without a model response. Contract counts normally stopped responses that fail the exact-array parser. Categories are disjoint and sum to Invalid. All are retained through the constant-3 fallback.}
\label{tab:format-failures}
\begin{tabular}{lllrrrr}
\toprule
Domain & Reader & Evidence & Invalid & Non-stop & Transport & Contract\\
\midrule
Coat & Qwen & full & 0 & 0 & 0 & 0\\
Coat & Qwen & memory & 2 & 0 & 0 & 2\\
Coat & Qwen & none & 1 & 0 & 0 & 1\\
Coat & Qwen & permuted & 0 & 0 & 0 & 0\\
Coat & Phi & full & 15 & 10 & 0 & 5\\
Coat & Phi & memory & 16 & 9 & 0 & 7\\
Coat & Phi & none & 23 & 0 & 0 & 23\\
Coat & Phi & permuted & 20 & 14 & 0 & 6\\
MovieLens & Qwen & full & 5 & 0 & 0 & 5\\
MovieLens & Qwen & none & 3 & 0 & 0 & 3\\
MovieLens & Qwen & permuted & 4 & 0 & 0 & 4\\
MovieLens & Phi & full & 1 & 1 & 0 & 0\\
MovieLens & Phi & none & 59 & 0 & 0 & 59\\
MovieLens & Phi & permuted & 1 & 1 & 0 & 0\\
\bottomrule
\end{tabular}
\end{table*}

\begin{table*}[t]
\centering\small
\caption{Post-freeze numerical precision sensitivity, specified before accuracy inspection. Columns show mean pairwise concordance from the original normal-equation solve, augmented least squares, and a common tie tolerance of 1e-10. Disagreements count users whose strict concordance differs between solvers. Primary MAE comparisons are unchanged.}
\label{tab:precision}
\begin{tabular}{llrrrrr}
\toprule
Domain & System & Users & Normal & Augmented & Near-tie & Disagreements\\
\midrule
Coat & History-only ridge & 199 & 0.637462 & 0.637287 & 0.637462 & 17\\
Coat & Ridge / permuted & 199 & 0.481885 & 0.481725 & 0.481859 & 24\\
MovieLens & History-only ridge & 200 & 0.557539 & 0.557547 & 0.557539 & 2\\
MovieLens & Ridge / permuted & 200 & 0.501437 & 0.501458 & 0.501437 & 3\\
\bottomrule
\end{tabular}
\end{table*}

\section{Interpretation Boundaries}
\label{app:limits}
\paragraph{Association use is not association correctness in general.} The intervention measures downstream usefulness of the observed assignments under one reader and task. Histories are finite and noisy. Metadata omit latent item features. A genuine association can be unhelpful for a particular target sample, and a model can respond to the intervention while worsening its predictions. Conversely, a null average effect can conceal opposing user-level effects. Reporting paired effect sizes and user-level variation avoids turning a control into an unsupported binary test of ``understanding.''

\paragraph{Preference ordering remains conditional on the dataset.} Within-user target ranking removes constant score shifts but does not remove item-quality effects. A model that predicts a generally popular item above an unpopular one can achieve useful concordance without learning an unusual individual preference. Our pairwise diagnostic is therefore a complement to the rating-assignment intervention, not an identification result for pure individual taste. Shared-item user comparisons such as PerRecBench address a different aspect of this problem.

\paragraph{Construction cost is not lifetime cost.} Writing is paid once per supplied history, while reading can recur. Token totals are counted separately for these stages. Memory-text bytes exclude embedding vectors, index files, database overhead, and source storage. No claim about universal dollar savings or isolated latency follows from a local sequential run. Reuse can amortize construction cost, but a break-even calculation requires an application-specific number of reads and model prices.

\paragraph{Numerical near-ties.} A pre-analysis check on fitting users found that two algebraically equivalent ridge solvers can produce predictions differing by less than $10^{-14}$ while disagreeing on an exact floating-point ranking tie. The primary MAE analysis is unchanged. The independent checker uses augmented least squares for continuous metrics and also reconstructs the frozen normal-equation arithmetic for exact-tie concordance. An auxiliary post-freeze sensitivity treats prediction differences at most $10^{-10}$ as ties and requires the two solvers to agree under that rule. Exact and tolerance-based results are retained, rather than conflating tiny numerical drift with a substantive ranking difference.

\paragraph{Bootstrap endpoint stability.} After the initial results and internal review, an auxiliary calculation repeated every primary operational contrast with 200,000 paired-user resamples for each of two fixed seeds, 20260921 and 20260922. All ordinary 95\% and family-adjusted 99.5\% zero-exclusion conclusions match the original analysis. The largest endpoint deviations from the frozen intervals are below 0.00196 and 0.00340 MAE, respectively. This assesses Monte Carlo sensitivity on the same observations, not independent data or an alternative primary analysis. The original 10,000-resample intervals remain the reported primary results.

\paragraph{Validation is computational and sample-based.} The held-out users were not used for model, prompt, or hyperparameter selection. The separate calculation check re-reads the archives, parses predictions independently, and recomputes metrics with a second implementation. It does not provide independent authorship, an external lab replication, or peer review. The two fixed model conversions and two small metadata tasks limit external validity. A wider memory-system ranking, longitudinal user study, or claim about conversational usefulness requires additional evidence and is outside this paper.

\end{document}